\documentclass[10pt,twocolumn,letterpaper]{article}
\usepackage{cvpr}
\usepackage{times}
\usepackage{epsfig}
\usepackage{graphicx}
\usepackage{amsmath}
\usepackage{amssymb}
\usepackage{color}
\usepackage{array}
\usepackage{float}
\usepackage{textcomp}
\usepackage{multirow}
\usepackage{amsmath}
\usepackage{amssymb}
\usepackage{mathrsfs}
\usepackage{commath}
\usepackage{algorithm2e}
\usepackage{caption} 
\usepackage[section]{placeins}
\usepackage[breaklinks=true,bookmarks=false]{hyperref}

\cvprfinalcopy % *** Uncomment this line for the final submission

\def\cvprPaperID{****} % *** Enter the CVPR Paper ID here
\def\httilde{\mbox{\tt\raisebox{-.5ex}{\symbol{126}}}}

\begin{document}

%% Because html converters don't know tabularnewline
\providecommand{\tabularnewline}{\\}
\floatstyle{ruled}
\newfloat{algorithm}{tbp}{loa}
\providecommand{\algorithmname}{Algorithm}
\floatname{algorithm}{\protect\algorithmname}

%%%%%%%%% TITLE
\title{Data-Driven Neuron Allocation for Scale Aggregation Networks}

\author{Yi Li \and Zhanghui Kuang \and Yimin Chen \and Wayne Zhang \and
SenseTime\\
{\tt\small \{liyi,kuangzhanghui,chenyimin,wayne.zhang\}@sensetime.com}
}

\maketitle
%\thispagestyle{empty}

%%%%%%%%% ABSTRACT
\begin{abstract}
Successful visual recognition networks benefit from aggregating information spanning from a wide range of scales. Previous research has investigated information fusion of connected layers or multiple branches in a block, seeking to strengthen the power of multi-scale representations. Despite their great successes, existing practices often allocate the neurons for each scale manually, and keep the same ratio in all aggregation blocks of an entire network, rendering suboptimal performance.
In this paper, we propose to learn the neuron allocation for aggregating multi-scale information in different building blocks of a deep network. The most informative output neurons in each block are preserved while others are discarded, and thus neurons for multiple scales are competitively and adaptively allocated. Our scale aggregation network (ScaleNet) is constructed by repeating a scale aggregation (SA) block that concatenates feature maps at a wide range of scales. Feature maps for each scale are generated by a stack of downsampling, convolution and upsampling operations. The data-driven neuron allocation and SA block achieve strong representational power at the cost of considerably low computational complexity. The proposed ScaleNet, by replacing all $3\times 3$ convolutions in ResNet with our SA blocks, achieves better performance than ResNet and its outstanding variants like ResNeXt and SE-ResNet, in the same computational complexity.
On ImageNet classification, ScaleNets absolutely reduce the top-1 error rate of ResNets by
1.12 (101 layers) and 1.82 (50 layers). On COCO object detection, ScaleNets absolutely improve the mmAP with backbone of ResNets by 3.6 (101 layers) and 4.6 (50 layers) on Faster RCNN, respectively. Code and models are released at \href{https://github.com/Eli-YiLi/ScaleNet}{https://github.com/Eli-YiLi/ScaleNet}.

%Visual recognition requires rich representations that span
%from small scale to large scale. A broad range of prior research has investigated
%information fusion of layers with different depths or convolutional kernel sizes, seeking to strengthen the multi-scale representational %power of a CNN.  In this paper, we instead propose a Scale Aggregation (SA) block by concatenating multi-scale feature maps, which are generated by a stack of downsampling, convolution and upsampling operations. The channel number of each scale is automatically learned by selecting most informative neurons. The block obtains representations with multi receptive fields. We show that SA blocks can replace existing convolutional layers (\eg, all $3\times 3$ layers) to form SANet architectures that performs extremely effectively on image classification and object detection. The proposed SANet with backbone ResNet101 achieves a top-1 error rate of 20.97\% on ImageNet and an mAP of 40.0 on COCO, absolutely reducing the top-1 error rate 2.63, and absolutely improving the
%mAP 3.3 compared with its baseline ResNet-101 while keeping the same computational complexity.
\end{abstract}
%%%%%%%%% BODY TEXT

\section{Introduction}

\begin{figure}
\setlength{\abovecaptionskip}{4pt}
\label{fig:fig_sa}
\centering
\includegraphics[width=0.45\textwidth]{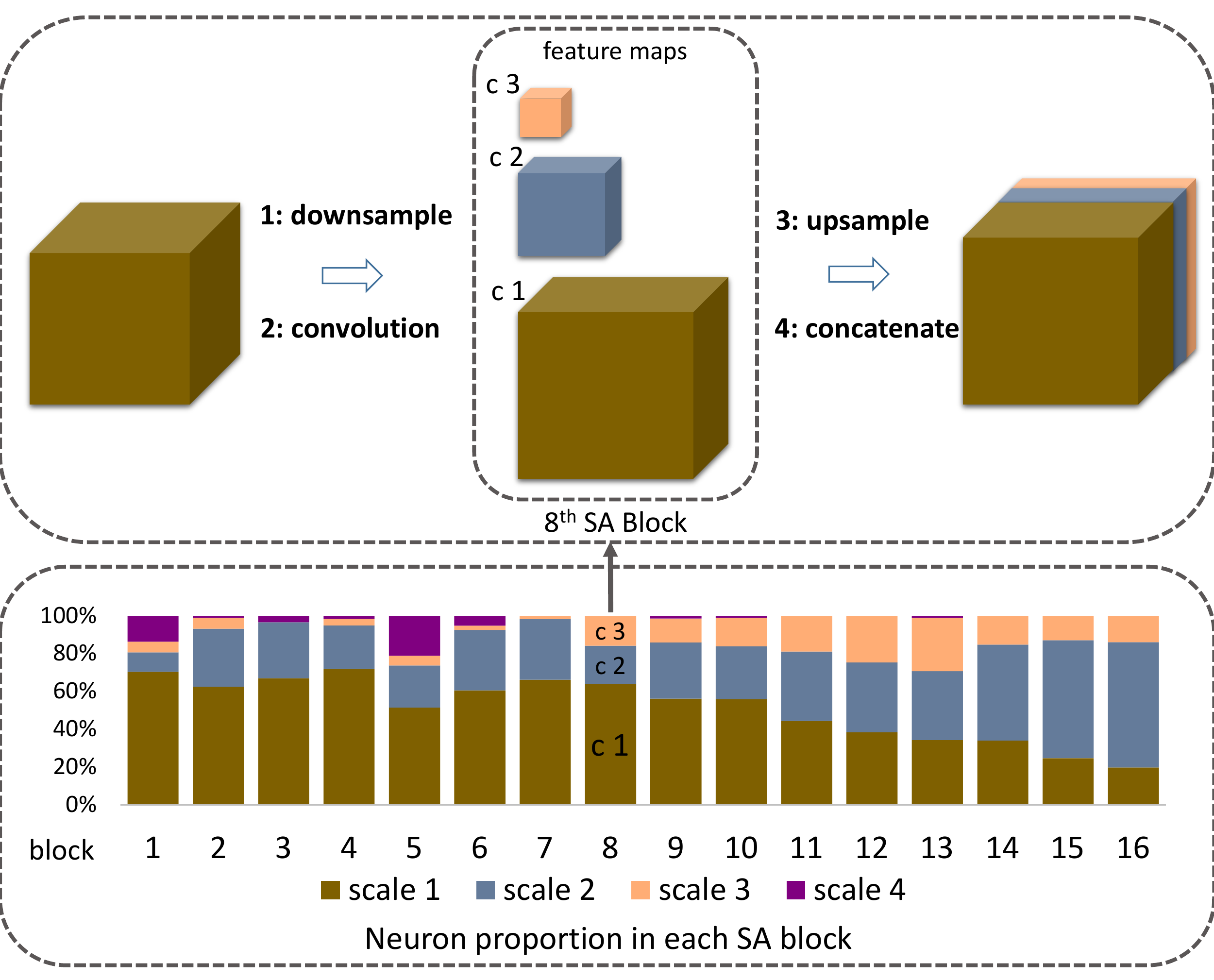}
\caption{Illustration of the data-driven neuron allocation for  the scale aggregation (SA) block. The proportion of output neurons (or channels) of different scales in an SA block is learned, and thus adaptively changes across layers in a network.}
\end{figure}

Deep convolutional neural networks (CNNs) have been successfully applied to a wide range of computer vision tasks,
such as image classification~\cite{krizhevsky2012imagenet}, object detection~\cite{Ren2017}, and semantic segmentation~\cite{Long2015}, due to their powerful end-to-end learnable representations. From bottom to top, the layers of CNNs have larger receptive fields with coarser scales, and their corresponding representations become more semantic. Aggregating context information from multiple scales has been proved to be effective for improving accuracy~\cite{Yu2018,he2016deep,huang2016densely,Lin2017a,Chen2018c,Bell2015,Kong2016,Christian_cvpr2015,Cao2018}. Small scale representations encode local structures such as textures, corners and edges, and are useful for localization, while coarse scale representations encode global contexts such as object categories, object interaction and scene, and thus clarify local confusion.

\begin{figure*}
\setlength{\abovecaptionskip}{4pt}
\begin{minipage}{0.5\linewidth}
\centerline{ \includegraphics[width =1\textwidth]{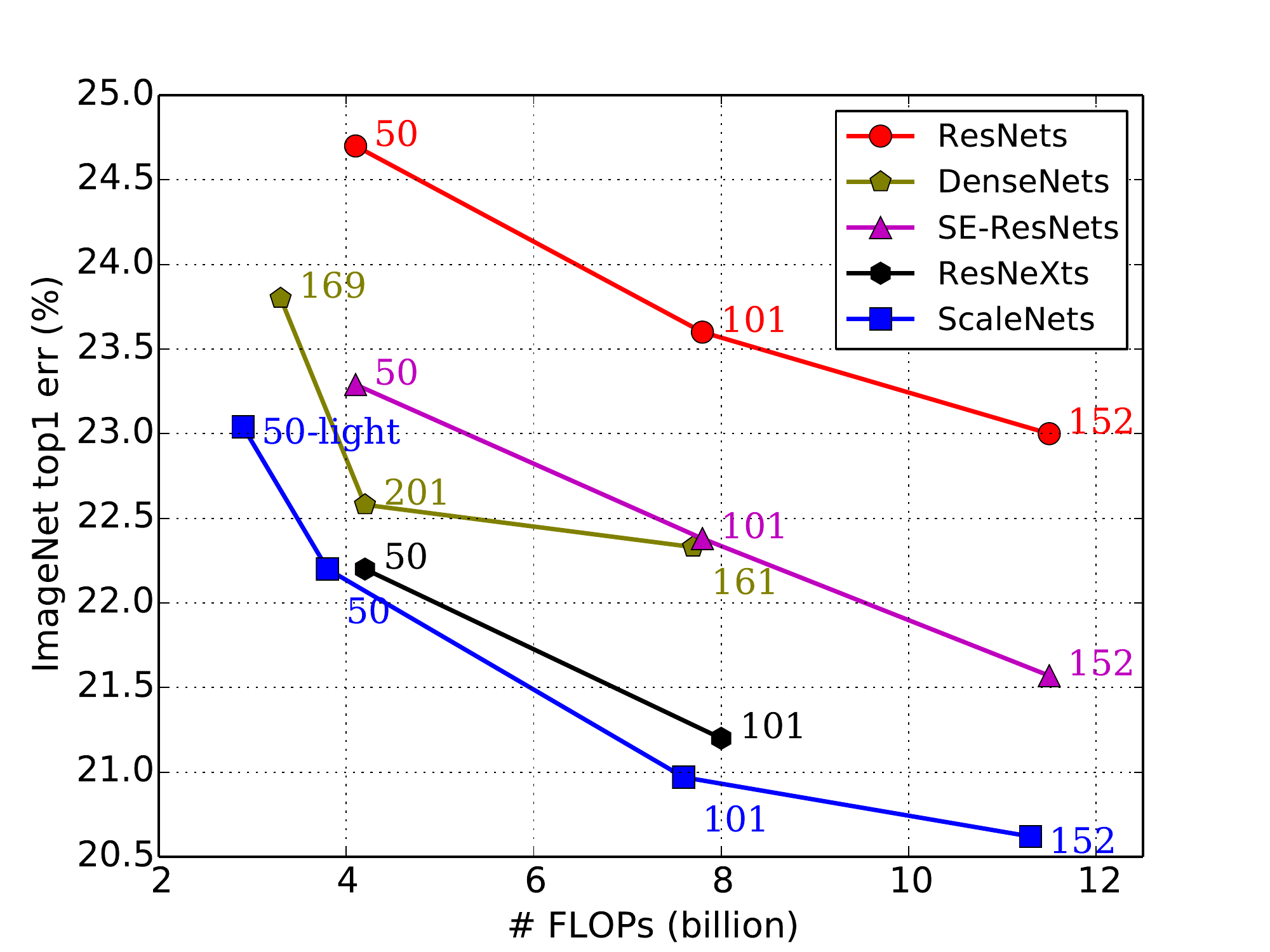}}
%\subcaption{  \label{figa} }
\end{minipage}
\begin{minipage}{0.5\linewidth}
\centerline{ \includegraphics[width =1\textwidth]{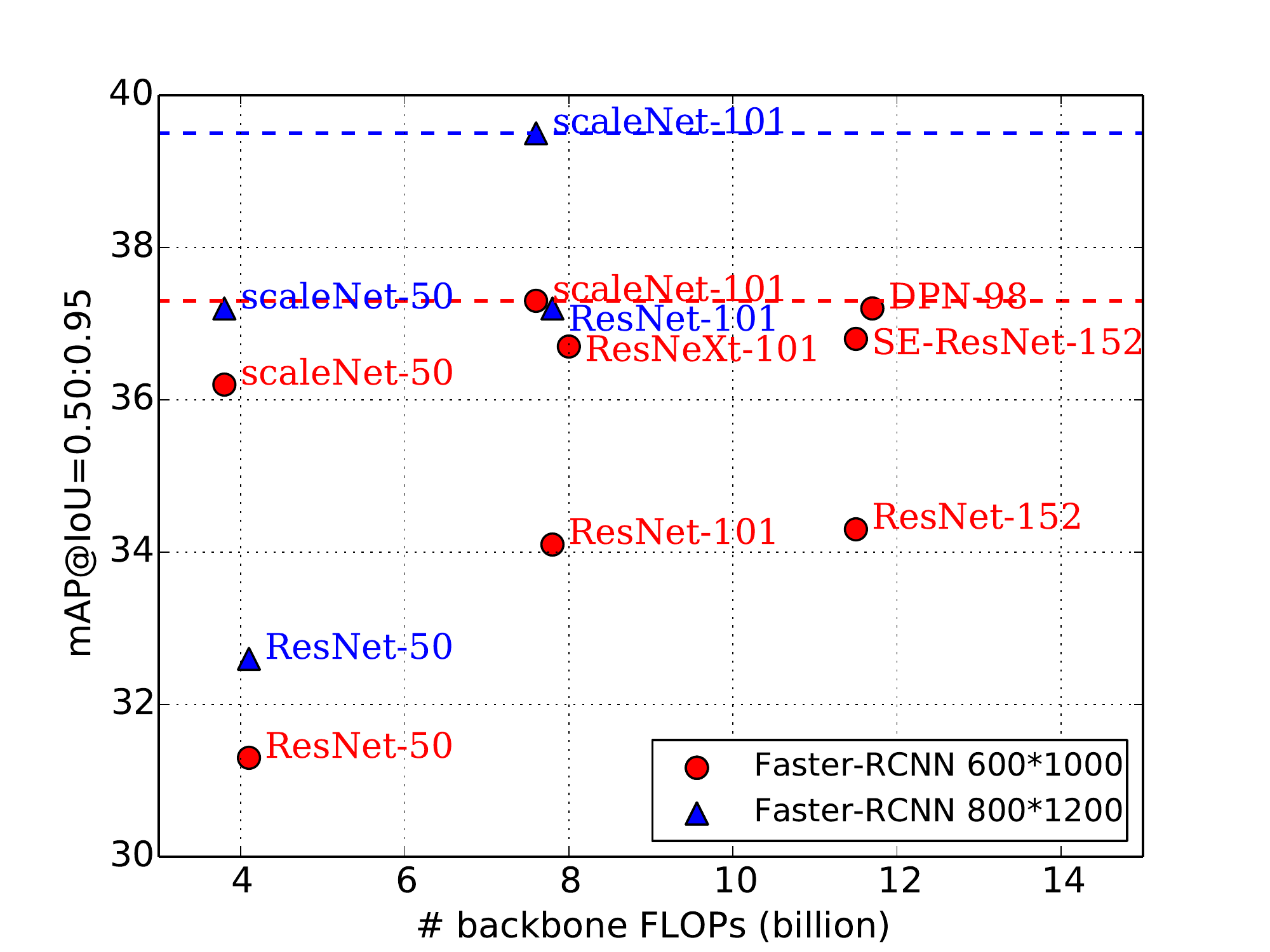}}
%\subcaption{  \label{figa} }
\end{minipage}
\captionsetup{margin=10pt}
\caption{Comparison of the ScaleNets and modern architectures' top-1 error rates (single-crop testing) on the ImageNet validation dataset (left) and mAP on MS COCO  mini-validation set (right) as a function of FLOPs during testing. ScaleNet-50-light indicates a light ScaleNet which is also constructed from ResNet-50. Architectures are given in the Appendix D.  }
\label{fig:fig_comp_performance}
\end{figure*}

There exist many previous attempts to fuse multi-scale representations by designing network architecture. They aggregate multi-scale representations of connected layers with different depths~\cite{Yu2018,he2016deep,huang2016densely,Lin2017a,Chen2018c,Bell2015,Kong2016,huang2018multiscale,NIPS2016_6304} or multiple branches in a block with different convolutional kernel sizes~\cite{Christian_cvpr2015,Cao2018}. The proportion of multi-scale representations for in each aggregation block is manually set in a trial-and-error process and kept the same in the entire network.  Ideally, the most efficient architecture design of multi-scale information aggregation is adaptive. The proportion of neurons for each scale is determinate according to the importance of the scale in gathering context. Such proportion should also be adaptive to the stage in the network. Bottom layers may prefer fine scales and top layers may prefer coarse scales.

%Inception~\cite{Christian_cvpr2015} concatenates multi scale representations, where each scale consists of different depth of convolutional layers with varied kernel sizes. Deep layer aggregation~\cite{Yu2018} iteratively and hierarchically merge the feature maps across layers.
%ResNet~\cite{he2016deep} and DenseNet~\cite{huang2016densely} can be considered as aggregating two or multiple layers with different receptive fields with skip connections. Multi-scale representation fusion is also widely employed in object detection by top-down or down-top across layer  connections~\cite{Lin2017a,Chen2018c,Bell2015,Kong2016}. Despite these methods, we note that aggregating multi-scale representations generated by reducing spatial resolution with different factors, while being a promising direction, is rarely studied before.

In this paper, we propose a novel data-driven neuron allocation method for multi-scale aggregation, which automatically learns the neuron proportion for each scale in all aggregation blocks of one network. We model the neuron allocation as one network optimization problem under FLOPs constraints which is solved by SGD and back projection. Concretely, we train one seed network with abundant output neurons for all scales using SGD, and then project the trained network into one feasible network that meets the constraints by selecting the top most informative output neurons amongst all scales. In this way, the neuron allocation for multi-scale representations is learnable and tailored for the network architecture.

To effectively extract and utilize multi-scale information, we present a simple yet effective Scale Aggregation (SA) block to strengthen the multi-scale representational power of CNNs.
Instead of generating multi-scale representations with connected layers of different depths or multi-branch different kernel sizes as done in~\cite{Christian_cvpr2015,he2016deep,Yu2018,he2016deep,huang2016densely,Lin2017a,Chen2018c,Bell2015,Kong2016}, an SA block explicitly downsamples the input feature maps with a group of factors to small sizes, and then independently conducts convolution, resulting in representations in different scales. Finally, the SA block upsamples the multi-scale representations back to the same resolution as that of the input feature maps and concatenate them in channel dimension together. We use SA blocks to replace all $3\times 3$ convolutions in ResNets to form ScaleNets.
Thanks to downsampling in each SA block, ScaleNets are very efficient by decreasing the sampling density in the spatial domain, which is independent yet complementary to network acceleration approaches in the channel domain. he proposed SA block is more computationally efficient and can capture a larger scale (or receptive field) range as shown in Figure ~\ref{fig:fig_rf_compare}, compared with previous multi-scale architecture.

We apply the proposed technique of data-driven neuron allocation to the SA block to form a learnable SA block. To demonstrate the effectiveness of the learnable SA block, we use learnable SA blocks to replace all $3\times3$ convolutions in ResNet to form a novel architecture called ScaleNet. The proposed ScaleNet outperforms ResNet and its outstanding variants such as ResNeXt~\cite{Xie2017} and SE-ResNet~\cite{Hu2018}, as well as recent popular architectures such as DenseNet~\cite{huang2016densely}, with impressive margins on image classification and object detection while keeping the same computational complexity as shown in Figure~\ref{fig:fig_comp_performance}. Specifically, ScaleNet-50 and ScaleNet-101 absolutely reduces the top-1 error rate of ResNet-101 and ResNet-50 by 1.12\% and 1.82\% on ImageNet respectively. Benefiting from the strong multi-scale representation power of learnable SA blocks, ScaleNets are considerably effective on object detection. The Faster-RCNN~\cite{Ren2017}  with backbone ScaleNet-101 and ScaleNet-50 absolutely improve the mmAP of those with ResNet-101 and ResNet-50 by 3.6 and 4.6 on MS COCO.

\section{Related Work}

Multi-scale representation aggregation has been studied for a long time. It can be categorized into shortcut connection approaches and multi-branch approaches.

\textbf{Shortcut connection approaches.} Connected layers with different depths usually have different receptive fields, and thus multi-scale representations. Shortcut connections between layers not only maximize information flow to avoid vanishing gradient, but also strengthen multi-scale representation power of CNNs. ResNet~\cite{he2016deep}, DenseNet~\cite{huang2016densely}, and Highway Network~\cite{Srivastava2015} fuse multi-scale information by identity shortcut connections or gating function based ones.  Deep layer aggregation~\cite{Yu2018} further extends shortcut connection with trees that cross stages. In object detection, FPN~\cite{Lin2017a} fuses coarse scale representations to fine scale ones from top to down in one detector's header~\cite{Lin2017a}. ASIF~{\cite{Chen2018c}} merges multi-scale representations from $4$ layers both from top to down and from down to top. HyperNet~\cite{Kong2016} and ION~\cite{Bell2015} concatenate multi-scale features from different layers to make prediction. All the shortcut connection approaches focus on reusing fine scale representations from preceding layers or coarse scale ones from subsequent layers. Due to limited connection patterns between layers, the scale (or receptive filed) range is limited. Instead, the proposed approach generates a wide range scale of representations with a group of downsampling factors itself in each SA block.  Moreover, it is a general and standard module which can  replace any convolutional layer of existing networks, and be effectively used in various tasks such as image classification and object detection as validated in our experiments.

\textbf{Multi-branch approaches.}  The most influential multi-branch network is GoogleNet~\cite{Christian_cvpr2015}, where each branch is designed with different depths and convolutional kernel sizes. Its branches have varied receptive fields and multi-scale representations.
Similar multi-branch network is designed for crowd counting in~\cite{Cao2018}. Different from previous multi-branch approaches, the proposed SA block generates multi-scale representations by downsampling the input feature maps by different factors to expand the scale of representations. Again, it can generate representations with wider  scale range than ~\cite{Christian_cvpr2015,Cao2018}.   Downsampling is also used in the context module of PSPNet~\cite{Zhao2017} and ParseNet~\cite{Rabinovich2016}. However, the context module is only used in the network header while the proposed SA block is used in the whole backbone and thus more general. Moreover, the neuron proportion for each scale is manually set and fixed  in the context module while automatically learned and different from one SA block to another in one network.

Our data-driven neuron allocation method is also related to network pruning methods~\cite{Gordon,Alvarez2016,Williams1995,Lebedev2015a} or network architecture search methods~\cite{Zoph2017,Veniat2017}. However, our data-driven neuron allocation method targets at multi-scale representation aggregation but not the whole architecture design. It learns the neuron proportion for scales in each SA block separately.
In this way, the neuron allocation problem is greatly simplified, and easily optimized.  
%Instead of pruning
%In the channel pruning methods, MorphNet is simple and effective. It is a kind of resource constraint for the whole network, which aim
%to allocate the limited computing resource to the most needed layer. We adopt this idea in single layer to allocate channels to appropriate
%scales and aviod the change in overall structure. Differently we maintain the amount of computation by reducing rather than expanding to %make sure the enough search scope of scales.
\begin{figure}
\setlength{\abovecaptionskip}{4pt}
\centering
\includegraphics[width=0.44\textwidth]{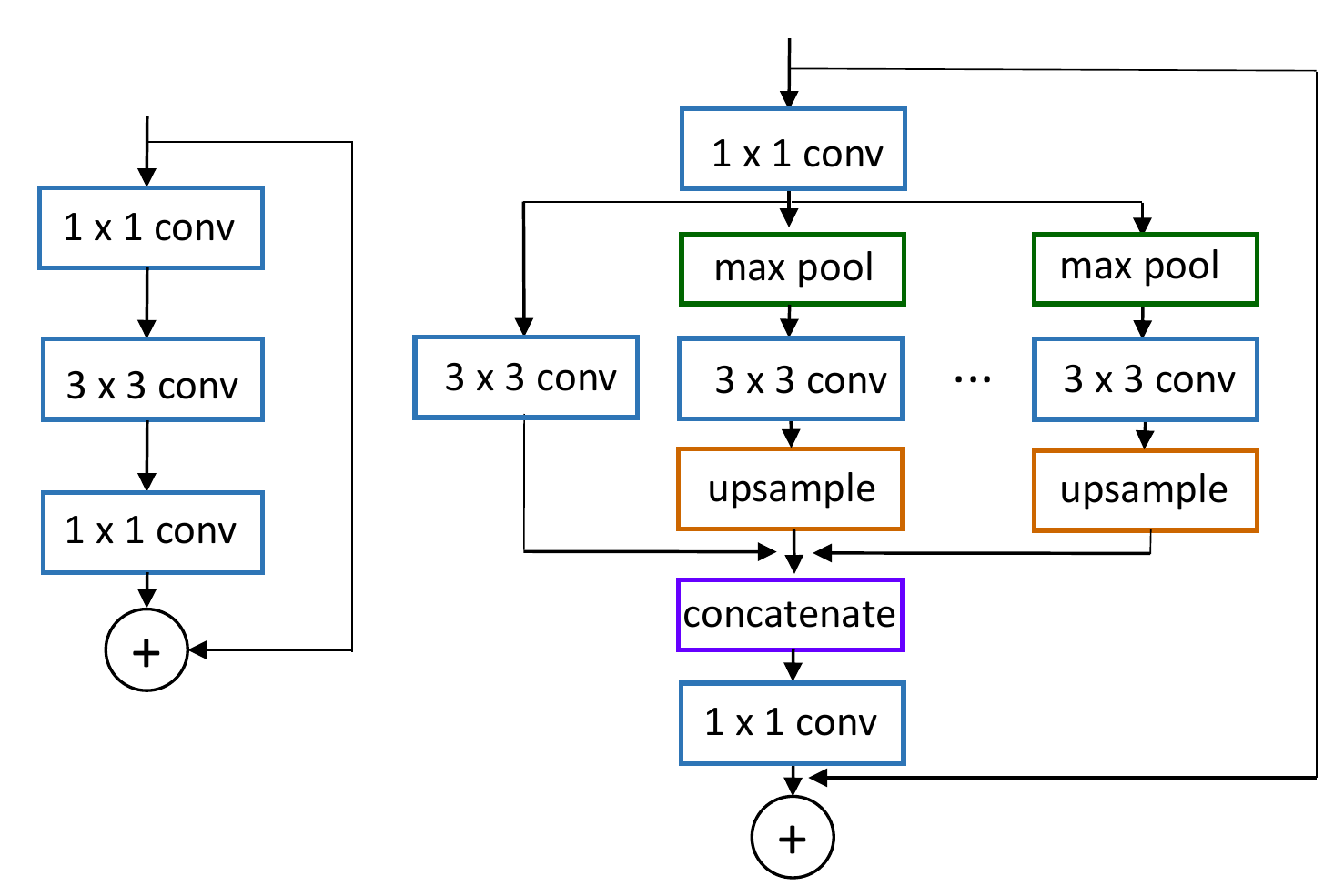}
\caption{Illustration of the SA block. The left shows the original residual block, and the right shows the module after replacing the $3\times 3$ convolution by the SA block. }
\label{fig:fig_conv3x3}
\end{figure}

\section{ScaleNets}
\subsection{Scale Aggregation Block}
The proposed scale aggregation block is a standard computational module which readily replaces any given transformation $\mathbf{Y}=\mathbf{T}(\mathbf{X})$, where $\mathbf{X}\in \mathbb{R}^{H\times W\times C}$, $\mathbf{Y}\in \mathbb{R}^{H\times W\times C_o}$ with $C$ and $C_o$ being the input and output channel number respectively. $\mathbf{T}$ is any operator such as a convolution layer or a series of convolution layers. Assume we have $L$ scales. Each scale $l$ is generated by sequentially conducting a downsampling $\mathbf{D}_l$, a transformation $\mathbf{T}_l$ and an unsampling operator $\mathbf{U}_l$:
\begin{equation}
\mathbf{X}^{'}_l=\mathbf{D}_l(\mathbf{X}),
\label{eq:eq_d}
\end{equation}
\begin{equation}
\mathbf{Y}^{'}_l=\mathbf{T}_l(\mathbf{X}^{'}_l),
\label{eq:eq_tl}
\end{equation}
\begin{equation}
\mathbf{Y}_l=\mathbf{U}_l(\mathbf{Y}^{'}_l),
\label{eq:eq_u}
\end{equation}
where $\mathbf{X}^{'}_l\in \mathbb{R}^{H_l\times W_l\times C}$,
$\mathbf{Y}^{'}_l\in \mathbb{R}^{H_l\times W_l\times C_l}$, and
$\mathbf{Y}_l\in \mathbb{R}^{H\times W\times C_l}$.
Notably, $\mathbf{T}_l$ has the similar structure as $\mathbf{T}$.
Substitute Equation~(\ref{eq:eq_d}) and~(\ref{eq:eq_tl}) into Equation~(\ref{eq:eq_u}), and concatenate all $L$ scales together, getting
\begin{equation}
\mathbf{Y}^{'}=\Vert^L_1\mathbf{U}_l(\mathbf{T}_l(\mathbf{D}_l(\mathbf{X}))),
\label{eq:eq_all}
\end{equation}
where $\Vert$ indicates concatenating feature maps along the channel dimension, and $\mathbf{Y}^{'} \in \mathbb{R}^{H\times W\times \sum^L_1 C_l}$ is the final output feature maps of the scale aggregation block.

In our implementation, the downsampling $\mathbf{D}_l$ with factor $s$ is implemented by a max pool layer with $s\times s$ kernel size and  $s$ stride. The upsampling $\mathbf{U}_l$ is implemented by resizing with the nearest neighbor  interpolation.

\subsection{Data-Driven Neuron Allocation}
There exist $L$ scales in each SA block. Different scales should
play different roles in blocks with different depths.
Therefore, simply allocating the output neuron proportion of scales equally would lead to suboptimal performance.
Our core idea is to identify the importance of each output neuron, and then
prune the unimportant neurons while preserving the important ones. 
For each output neuron, we employ its corresponding  scale weight ($\gamma$) of its subsequent BatchNorm~\cite{ioffe2015batch} layer to evaluate its importance. The underlying reason is that $\gamma$ is positively correlated with the output confidence of its corresponding neuron.

Let $K$, $O_k$ ($1\leq k\leq K$), and  $O_{kl}$ ($1\leq k\leq K$ and $1\leq l\leq L$)  denote the total SA block index of the target network,  the computational complexity budget of the $k^{th}$ SA block, and the computational complexity of one output neuron at scale $l$ in the $k^{th}$ block respectively. We target at optimally allocating neurons for each scale in the $k^{th}$ SA block  with the budget $O_k$.  Formally, we have
\begin{equation}
\min\limits_{\theta}\ F(\theta), \ \text{s.t.} \ \forall k, \sum_{1\leq n \leq N_k } O_{kl(\theta_{kn})} \leq O_k,
\label{eq:eq_loss}  
\end{equation}
where $F(\theta)$ is the loss function of the whole network with $\theta$ being the learnable weights of the network, and $\theta_{kn}$ being the weight of $n^{th}$ output neuron in the $k^{th}$ SA block. $l(\cdot)$ indicates the scale index, and $N_k$ is the total number of
output neurons in the $k^{th}$ SA block.

We optimize the objective function~(\ref{eq:eq_loss}) by SGD with projection.
We first optimize $F(\theta)$, getting $\theta^t$, we then project $\theta_t$ back to the feasible domain defined by constraints for each SA block $k$ by optimizing
\begin{align}
\begin{split}
\min\limits_{\theta}& \sum\limits_{n} \abs{\mathbf{V}(\theta_{kn})-\mathbf{V}(\theta_{kn}^t)}\\
&\text{s.t.}\ \sum_{1\leq n \leq N_k } O_{kl(\theta_{kn})} \leq O_k,
\end{split}
\label{eq:eq_back}
\end{align}
where $\mathbf{V}(\theta_{kn})$ indicates the importance of the $n^{th}$ neuron in the  $k^{th}$ SA block. It is defined to be the scale weight  which corresponds to the target channel $(k,n)$  in its subsequent BatchNorm layer as done in \cite{Gordon}. The more important and of less computational complexity the neuron is, the more likely it should be preserved. Equation~(\ref{eq:eq_back}) is greedily solved by selecting neurons with top biggest $\mathbf{V}(\theta_{kn})/O_{kl(\theta_{kn})}^b $ in the  $k^{th}$ SA block. Note that $b$ is an exponential balance factor of computational complexity. We found $b=0$ achieves good results in our experiments.

Algorithm~\ref{alg:alg_1} lists the procedure of neuron allocation.
First, we set a seed network by setting $C$ neurons for each scale (\ie, $N_k=CL$).
Second, we train the seed network till convergence.
Third, we select the top most important neurons in SA blocks by solving Equation(~\ref{eq:eq_back}), and get a new network.
Finally, we retrain the new network from scratch.
\begin{algorithm}
\caption{Data-driven neuron allocation}
\label{alg:alg_1}
Initialize a seed network by setting $N_k=LC$\\
Train the seed network till convergence\\
\For{$k=1:K$ }{
\For{$n=1:N_k$}{Compute $p_{kn}=\mathbf{V}(\theta_{kn})/O_{kl(\theta_{kn})}^b $}
Select neurons with top biggest $p_{kn}$ under the constraint of Equation~(\ref{eq:eq_back})
\;}
Retrain the new network till convergence.
\end{algorithm}

\subsection{Instantiations}
\label{sec:sec_inst}
The proposed SA block can be integrated into standard architectures by replacing its existing convolutional layers or modules. To illustrate this point, we develop ScaleNets by incorporating SA blocks into the recent popular ResNets~\cite{he2016deep}.

In ResNets~\cite{he2016deep}, $3\times 3$ convolutions account for most of the whole network computational complexity. Therefore, we replace all $3\times 3$ layers with SA blocks as shown in Figure~\ref{fig:fig_conv3x3}. We replace the stride in $3\times 3$ convolution by extra max pool layer as done in DenseNets~\cite{huang2016densely}.
In this way, all $3\times 3$ layers can be replaced by SA blocks consistently. As shown in Table~\ref{tab:tab_archetecture}, using ResNet-50, ResNet-101, and ResNet-152 as the start points, we obtain the corresponding ScaleNets\footnote{Note that the number indicates the layer number of their start points but not ScaleNets.} by setting the computational complexity budget of each SA block to that of its corresponding $3\times 3$ conv in the residual block during the neuron allocation procedure.

\begin{table*}
\setlength{\abovecaptionskip}{4pt}
\begin{tabular}{c|c|c|c}
\hline
output size & ScaleNet-50 & ScaleNet-101 & ScaleNet-152\tabularnewline
\hline
112\texttimes 112 & \multicolumn{3}{c}{$7\times 7$ conv, stride 2}\tabularnewline
\hline
56\texttimes 56 & \multicolumn{3}{c}{$3\times 3$ max pool, stride 2}\tabularnewline
\hline
56\texttimes 56 & $\left[\begin{array}{c}
\text{$1\times 1$ conv},64\\
\mathbf{D}_{[1,2,4,7]}\\
\text{$3\times 3$ conv}_{[C_l,C_2,C_3,C_4]}\\
\mathbf{U}_{[1,2,4,7]}\\
\text{$1\times 1$ conv},256
\end{array}\right]$\texttimes 3 & $\left[\begin{array}{c}
\text{$1\times 1$ conv},64\\
\mathbf{D}_{[1,2,4,7]}\\
\text{$3\times 3$ conv}_{[C_1,C_2,C_3,C_4]}\\
\mathbf{U}_{[1,2,4,7]}\\
\text{$1\times 1$ conv},256
\end{array}\right]$\texttimes 3 & $\left[\begin{array}{c}
\text{$1\times 1$ conv},64\\
\mathbf{D}_{[1,2,4,7]}\\
\text{$3\times 3$ conv}_{[C_1,C_2,C_3,C_4]}\\
\mathbf{U}_{[1,2,4,7]}\\
\text{$1\times 1$ conv},256
\end{array}\right]$\texttimes 3\tabularnewline
\hline
28\texttimes 28 & \multicolumn{3}{c}{\text{$2\times 2$ max pool, stride 2}}\tabularnewline
\hline
28\texttimes 28 & $\left[\begin{array}{c}
\text{$1\times 1$ conv},128\\
\mathbf{D}_{[1,2,4,7]}\\
\text{$3\times 3$ conv}_{[C_1,C_2,C_3,C_4]}\\
\mathbf{U}_{[1,2,4,7]}\\
\text{$1\times 1$ conv},512
\end{array}\right]$\texttimes 4 & $\left[\begin{array}{c}
\text{$1\times 1$ conv},128\\
\mathbf{D}_{[1,2,4,7]}\\
\text{$3\times 3$ conv}_{[C_1,C_2,C_3,C_4]}\\
\mathbf{U}_{[1,2,4,7]}\\
\text{$1\times 1$ conv},512
\end{array}\right]$\texttimes 4 & $\left[\begin{array}{c}
\text{$1\times 1$ conv},128\\
\mathbf{D}_{[1,2,4,7]}\\
\text{$3\times 3$ conv}_{[C_1,C_2,C_3,C_4]}\\
\mathbf{U}_{[1,2,4,7]}\\
\text{$1\times 1$ conv},512
\end{array}\right]$\texttimes 8\tabularnewline
\hline
14\texttimes 14 & \multicolumn{3}{c}{\text{$2\times 2$ max pool, stride 2}}\tabularnewline
\hline
14\texttimes 14 & $\left[\begin{array}{c}
\text{$1\times 1$ conv},256\\
\mathbf{D}_{[1,2,4,7]}\\
\text{$3\times 3$ conv}_{[C_1,C_2,C_3,C_4]}\\
\mathbf{U}_{[1,2,4,7]}\\
\text{$1\times 1$ conv},1024
\end{array}\right]$\texttimes 6 & $\left[\begin{array}{c}
\text{$1\times 1$ conv},256\\
\mathbf{D}_{[1,2,4,7]}\\
\text{$3\times 3$ conv}_{[C_1,C_2,C_3,C_4]}\\
\mathbf{U}_{[1,2,4,7]}\\
\text{$1\times 1$ conv},1024
\end{array}\right]$\texttimes 23 & $\left[\begin{array}{c}
\text{$1\times 1$ conv},256\\
\mathbf{D}_{[1,2,4,7]}\\
\text{$3\times 3$ conv}_{[C_1,C_2,C_3,C_4]}\\
\mathbf{U}_{[1,2,4,7]}\\
\text{$1\times 1$ conv},1024
\end{array}\right]$\texttimes 36\tabularnewline
\hline
$7\times 7$ & \multicolumn{3}{c}{\text{$2\times 2$ max pool, stride 2}}\tabularnewline
\hline
$7\times 7$ & $\left[\begin{array}{c}
\text{$1\times 1$ conv},512\\
\mathbf{D}_{[1,2,4,7]}\\
\text{$3\times 3$ conv}_{[C_1,C_2,C_3,C_4]}\\
\mathbf{U}_{[1,2,4,7]}\\
\text{$1\times 1$ conv},2048
\end{array}\right]$\texttimes 3 & $\left[\begin{array}{c}
\text{$1\times 1$ conv},512\\
\mathbf{D}_{[1,2,4,7]}\\
\text{$3\times 3$ conv}_{[C_1,C_2,C_3,C_4]}\\
\mathbf{U}_{[1,2,4,7]}\\
\text{$1\times 1$ conv},2048
\end{array}\right]$\texttimes 3 & $\left[\begin{array}{c}
\text{$1\times 1$ conv},512\\
\mathbf{D}_{[1,2,4,7]}\\
\text{$3\times 3$ conv}_{[C_1,C_2,C_3,C_4]}\\
\mathbf{U}_{[1,2,4,7]}\\
\text{$1\times 1$ conv},2048
\end{array}\right]$\texttimes 3\tabularnewline
\hline
$1\times 1$ & \multicolumn{3}{c}{\text{avg pool, 1000-d fc, softmax}}\tabularnewline
\hline
\end{tabular}

\caption{Architectures of ScaleNets. $\mathbf{D}_{[1,2,4,7]}$ indicates $1\times 1$, $2\times 2$, $4\times 4$, and $7\times 7$ downsampling layers. $\mathbf{U}_{[1,2,4,7]}$ indicates $1\times 1$, $2\times 2$, $4\times 4$, and $7\times 7$ upsampling layers. We select $7\times 7$ (but not 8\texttimes 8) downsampling and upsampling layers since the spatial resolution of last stage of networks  is $7\times 7$. $ \text{$3\times 3$ conv}_{[C_1,C_2,C_3,C_4]}$ indicates $3\times 3$ convolution layers with output channels of $C_1$, $C_2$, $C_3$, and $C_4$. Note that $C_1$, $C_2$, $C_3$, and $C_4$ are different from one SA block to another, and are detailed in the Appendix D.}
\label{tab:tab_archetecture}
\end{table*}

\label{sec:constraint}
\subsection{Computational Complexity}
The proposed SA block is of practical use. It makes ScaleNets efficient, because the feature maps are smaller. Theoretically, if we set the output channel number of one SA block to $C$ (\ie, $\sum^L_1 C_l=C$), the saved FLOPs is $9C(\sum^L_1H_lW_lC_l) - 9C(HWC)$. Taking ScaleNet-50-light as an example, it reduces FLOPs of its start point ResNet-50 by 29\%   while absolutely improving the single-crop top-1 accuracy by 0.98 on ImageNet as shown in Table~\ref{tab:light}. Also we compare the ScaleNet-50-light with the state-of-art pruning methods in Appendix B. Besides the FLOPs, we evaluate the real GPU time in Appendix C.

\begin{table}
\center
\small
\begin{tabular}{l|l|l|l}
\hline
 & top-1 err. & top-5 err. & GFLOPs\tabularnewline
\hline
ResNet-50 & $24.02$ & $7.13$ &  $4.1$\tabularnewline
ScaleNet-50-light & \textbf{23.04}$_{(-0.98)}$ & \textbf{6.66}$_{(-0.47)}$ & $\textbf{2.9}_{(-1.2)}$\tabularnewline
\hline
\end{tabular}
\caption{Efficiency of ScaleNet-light. All results are evaluated with single crop on ImageNet validation set. ScaleNet-50-light indicates a light ScaleNet constructed from ResNet-50. Results of ResNet-50 are reimplemented with the same training strategy as ScaleNets for fair comparison. }
\label{tab:light}
\end{table}

\subsection{Implementation}
Our implementation for ImageNet follows the practice in~\cite{he2016deep,Hu2018,Christian_cvpr2015}.
We perform standard data augmentation with random cropping, random horizontal flipping and photometric distortions~\cite{Christian_cvpr2015} during training. All input images are resized to
224\texttimes 224 before feeding them into networks. Optimization is performed using synchronous SGD with momentum 0.9, weight decay 0.0001 and batch size 256 on servers with 8 GPUs. The initial learning rate is set to 0.1 and decreased by a factor of 10 every 30 epoches. All models are trained for 100 epoches from scratch.

On CIFAR-100, we train models with a batch size of 64 for
300 epoches. The initial learning rate is set to 0.1, and is reduced
by 10 times in 150 and 225. The data augmentation only includes random
horizontal flipping and random cropping with 4 pixels padding.

On MS COCO, we train all detection models  using the publicly available implementation\footnote{https://github.com/jwyang/faster-rcnn.pytorch} of Faster RCNN.
Models are trained on servers with 8 GPUs. The batch size and epoch number are set to 16 and  10 respectively. The initial learning rate is set to 0.01 and reduced by a factor of 10 at epoch 4 and epoch 8.

\section{Experiments}
\subsection{ImageNet Classification}
We evaluate our method on the ImageNet 2012 classification dataset~\cite{krizhevsky2012imagenet} that consists of 1000 classes. The models are trained on the 1.28 million training images, and evaluated on the 50k validation images with both top-1 and top-5 error rate. When evaluating the models we apply centre-cropping so that 224\texttimes 224 pixels are cropped from each image after its shorter edge is first resized to 256.

\begin{table*}
\setlength{\abovecaptionskip}{4pt}
\center
\begin{tabular}{l|l|l|l|l|l|l|l|l}
\hline
method & \multicolumn{2}{c|}{original} & \multicolumn{3}{c|}{re-implementation} & \multicolumn{3}{c}{ScaleNet}\\
\cline{2-9}
 & top-1 err. & top-5 err.  & top-1 err. & top-5 err. & GFLOPs & top-1 err. & top-5 err. & GFLOPs\\ \hline
ResNet-50&24.7&7.8&24.02&7.13&4.1&\textbf{22.20}$_{(-1.82)}$&\textbf{6.04}$_{(-1.09)}$&\textbf{3.8} \\ \hline
ResNet-101&23.6&7.1&22.09&6.03&7.8&\textbf{20.97}$_{(-1.12)}$&\textbf{5.58}$_{(-0.45)}$&\textbf{7.5} \\ \hline
ResNet-152&23.0&6.7&21.58&5.75&11.5&\textbf{20.62}$_{(-0.96)}$&\textbf{5.34}$_{(-0.41)}$&\textbf{11.2} \\ \hline
\end{tabular}
\caption{Comparisons between ScaleNets and their baseline ResNets with single-crop error rates (\%) on ImageNet validation set.
The original column refers to the reported results in the original paper. For fair comparison, we retrain the baselines using the same strategy of training ScaleNet and report the results in the reimplementation column.} 
\label{tab:tab_com_baseline}
\end{table*}
\textbf{Comparisons with baselines.} We begin evaluations by comparing the proposed ScaleNets with their corresponding baseline networks in Table~\ref{tab:tab_com_baseline}. It has been shown that ScaleNets with different depths consistently improve their baselines with impressive margins while using comparable (or even a little less) computational complexity. Specifically, Compared with baselines, ScaleNet-50, 101, and 152 absolutely 
reduce the top-1 error rate by 1.82, 1.12 and 0.96, the top-5 error rate by 1.09, 0.45, and 0.41 on ImageNet respectively. ScaleNet-101 even outperforms ResNet-152, although it has only 66\% FLOPs (7.5 \vs 11.5). It suggests that explicitly and effectively aggregating multi-scale representations of ScaleNets can achieve considerably much performance gain on image classification although deep CNNs are robust against scale variance to some extent.
\begin{table}
\setlength{\abovecaptionskip}{4pt}
\center
\begin{tabular}{l|l|l|l}
\hline
method&top-1 err.& top-5 err.& GFLOPs \\ \hline
ResNeXt-50& 22.2&-&4.2 \\
ResNeXt-101& 21.2&5.6&8.0  \\ \hline
SE-ResNet-50&23.29&6.62&4.1 \\
SE-ResNet-101&22.38&6.07&7.8 \\
SE-ResNet-152&21.57&5.73&11.5 \\ \hline
DenseNet-121& 25.02&7.71&2.9\\
DenseNet-169& 23.8&6.85&3.4\\
DenseNet-201& 22.58&6.34&4.3\\ \hline
ScaleNet-50& \textbf{22.2}&\textbf{6.04}&\textbf{3.8} \\
ScaleNet-101& \textbf{20.97}&\textbf{5.58}&\textbf{7.5}  \\
ScaleNet-152& \textbf{20.62}&\textbf{5.34}&\textbf{11.2}  \\ \hline
\end{tabular}
\caption{Comparison with state-of-the-art architectures with single-crop top-1 and top-5 error rates (\%) on ImageNet validation set.}
\label{tab:tab_com_start_of_the_art}
\end{table}

\textbf{Comparisons with state-of-the-art architectures.} We next compare ScaleNets with ResNets, ResNeXts, SE-ResNets, and DenseNets in Table~\ref{tab:tab_com_start_of_the_art}. It has been shown that ScaleNets consistently outperform them. Remarkably, ScaleNet-50, 101 and 152 absolutely reduce the top-1 error rate by 1.09 , 1.41 and 0.95 compared with their counterparts SE-ResNet-50, 101 and 152 respectively. Surprisingly, our ScaleNets-101 performs better than ResNeXt-101 by 0.23  without group convolutions which are not GPU friendly. We also evaluate the running time in Appendix C, which suggests ScaleNet achieves the best accuracy on ImageNet while with the least GPU running time

%\textbf{Network acceleration.}
\subsection{CIFAR Classification}
We also conduct experiments on CIFAR-100 dataset~\cite{Krizhevsky2009}. 
To make full use of the same SA block architecture, 
our baseline ResNets on CIFAR-100 also employ residual bottleneck blocks (\ie, a subsequent layers of $1\times 1$ conv, $3\times 3$ conv and $1\times 1$ conv) instead of basic residual blocks (two $3\times 3$ conv layers) in~\cite{he2016deep}. The network inputs are $32\times 32$ images. The first layer is $3\time 3$ convolutions with 16 channels. Then we use a stack of $n$ residual bottleneck blocks on each of these three stages with the feature maps of sizes 32\texttimes 32, 16\texttimes 16 and 8\texttimes 8 respectively. The numbers of channels for $1\times 1$ conv , $3\times 3$  conv, and $1\times 1$ conv in each residual block are set to 16, 16 and 64 on the first stage, 32, 32 and 128 on the second stage, 64, 64 and 256 on the third stage. The subsampling is performed by convolutions with a stride of 2 at beginning of each stage. The network ends
with a global average pool layer, a 100-way fully-connected layer, and softmax layer. There are totally 9n+2 stacked weighted layers. When $n=4, 6,$ and $10$, we get baselines ResNet-38, ResNet-56 and ResNet-101 respectively for tiny images. Their corresponding ScaleNets with comparable computational complexity are denoted by ScaleNet-38, ScaleNet-56, and ScaleNet-101.

We compare the performances between ScaleNets and their baselines on CIFAR-100 in Table~\ref{tab:tab_cifar}. Again, the proposed ScaleNets outperforms ResNets with big margins. It has been validated that ScaleNets can effectively enhance and improve its strong baseline ResNets across multiple datasets from ImageNet to CIFAR-100, and multi-scale aggregation is also important for tiny image classification. 
%\par\end{flushleft}
\begin{table}
\setlength{\abovecaptionskip}{4pt}
\center
\begin{tabular}{l|l|l}
\hline
\# layer &  ResNets & ScaleNets\tabularnewline
\hline
38 layers  & $26.88$ & $\textbf{24.60}_{(-2.28)}$\tabularnewline
56 layers & $26.19$ & $\textbf{23.83}_{(-2.36)}$\tabularnewline
101 layers  & $24.54$ & $\textbf{22.77}_{(-1.77)}$\tabularnewline
\hline
\end{tabular}
\caption{Comparisons of the top-1 error rate  on CIFAR-100 between ScaleNets and their baseline ResNets. All the results are the best of 5 runs.}
\label{tab:tab_cifar}
\end{table}
\subsection{Data-Driven Neuron Allocation}

\begin{figure}
\setlength{\abovecaptionskip}{4pt}
\center
%\small
\begin{minipage}{0.8\linewidth}
\centerline{ \includegraphics[width =1\textwidth]{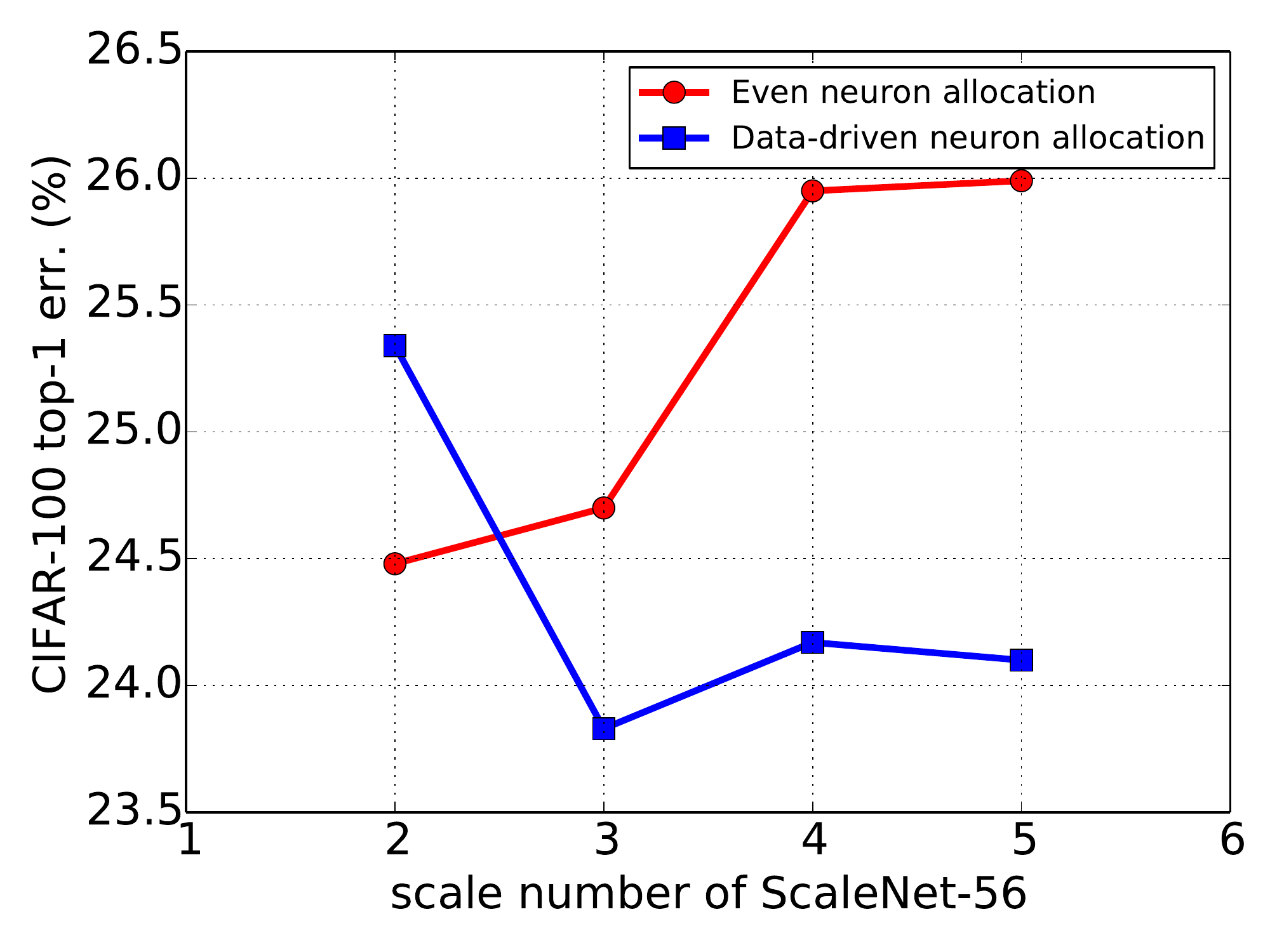}}
%\subcaption{  \label{figa} }
\end{minipage}
\begin{minipage}{0.8\linewidth}
\centerline{ \includegraphics[width =1\textwidth]{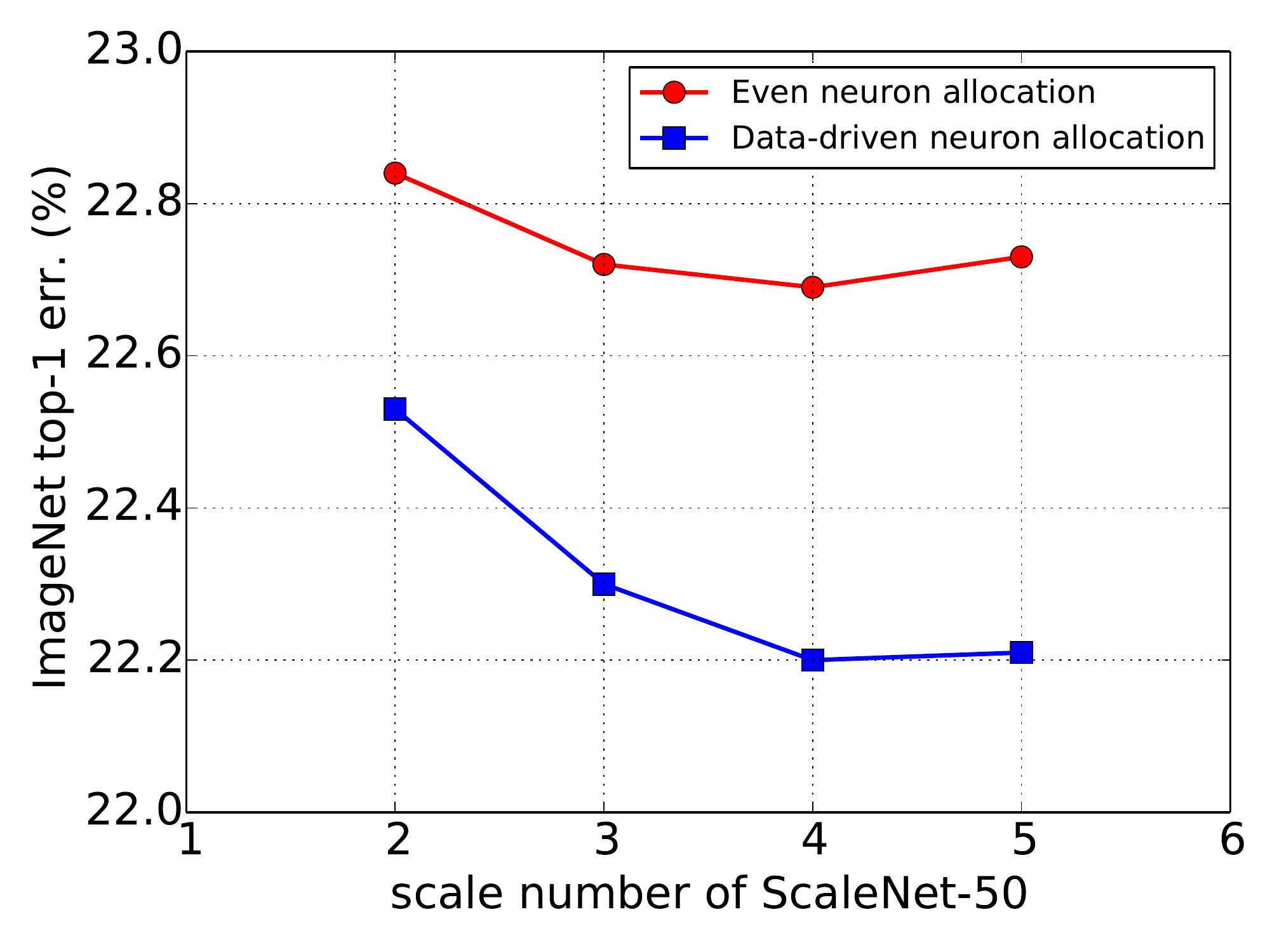}}
%\subcaption{  \label{figa} }
\end{minipage}
\caption{Comparisons between even neuron allocation and data-driven neuron allocation on CIFAR-100 and ImageNet.}
\label{fig:fig_comp_even}
\end{figure}

The proposed ScaleNets can automatically learn the neuron proportion for each scale in each SA block.
The neuron allocation depends on the training data distribution and network architectures.

\textbf{Even allocation \vs data-driven allocation.} Figure~\ref{fig:fig_comp_even} compares even neuron allocation for scales in each SA block and data-driven neuron allocation. We conduct experiments on both CIFAR-100 and ImageNet with scale number $L$ from $2$ to $5$. Data-driven neuron allocation outperforms even allocation with impressive margins in all setting except that on CIFAR-100 with $L=2$. 
We also observe that data-driven allocation performs best on CIFAR-100 with $L=3$ and ImageNet with $L=4$. This is reasonable since ImageNet has bigger resolution and needs representation with wider scale range than CIFAR-100.  We set $L$ to 3 on CIFAR-100 and $L$ to 4 on ImageNet in all our experiments except otherwise noted. Based on even allocation (gains from SA block), ScaleNet-50 achieves top-1 error rate of 22.76\%. With data-driven allocation, the top-1 error rate can be further reduced to 22.20\%.

\begin{figure}
\setlength{\abovecaptionskip}{4pt}
\centering
\tiny
\includegraphics[width=0.44\textwidth]{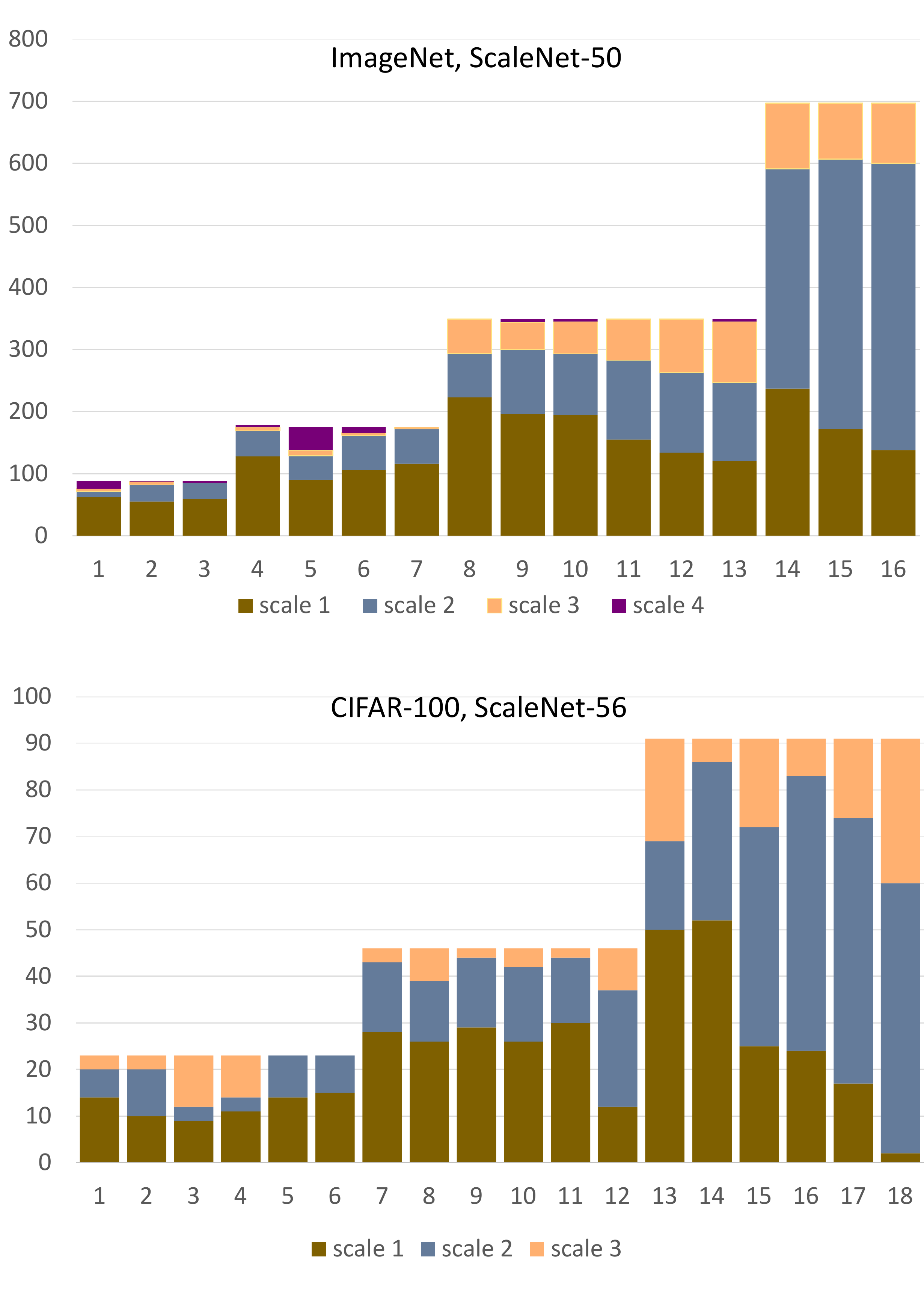}
\caption{Neuron proportion for scales in each SA block of ScaleNets on CIFAR-100 and ImageNet.}
\label{fig:fig_learned_scale}
\end{figure}

\textbf{Visualization of neuron allocation.} Figure~\ref{fig:fig_learned_scale} shows learned neuron proportion in each SA block of ScaleNets. We observe that neuron proportions for scales are different from one SA block to another in one network. 
Specifically, scale 2 accounts for more and more proportion from bottom to top on both CIFAR-100 and ImageNet. 
Scale 4 mainly exists in the first two stages of ScaleNet-50 on ImageNet. 
\subsection{Object Detection on MS COCO}
To further evaluate the generalization on other recognition tasks, we conduct object detection experiments
on MS COCO~\cite{Lin2014} consisting of 80k training images and 40k validation images, which are further split into 
 35k miniusmini and 5k mini-validation set. Following the common setting~\cite{he2016deep},
 we combine the training images and miniusmini images and thus obtain 115k images for training,  and the 5k mini-validation set for
 evaluation. We employ the Faster RCNN framework~\cite{Ren2017}. We test models by resizing the shorter edge of image to 800 (or 600) pixels, and restrict the max size of the longer edge to 1200 (or 1000).
 
\textbf{Comparisons with baselines.} Table~\ref{tab:tab_coco} compares the detection results of ScaleNets and their baseline ResNets on MS COCO. With multi-scale aggregation, Faster RCNN achieves impressive gains with range from $3.2$ to $4.9$. Especially, ScaleNet-101 reaches an mmAP of 39.5.

\begin{table*}
\setlength{\abovecaptionskip}{4pt}
\center
\begin{tabular}{l|l|l|l|l|l|l|l|l}
\hline
\multirow{2}{*}{} & \multicolumn{4}{c|}{600/1000} & \multicolumn{4}{c}{800/1200}\tabularnewline
\cline{2-9}
 & mmAP & AP$_s$ & AP$_m$ & AP$_l$ & mmAP & AP$_s$ & AP$_m$ & AP$_l$\tabularnewline
\hline
ResNet-50& 31.7&	12.6&	35.9&	48.3&	32.6&	15.9&	36.7&	46\tabularnewline
ScaleNet-50&36.2&	17.1&	40.7&	53.8&	37.2&	19.4&	41.3&	52.6\tabularnewline
\hline
ResNet-101&34.1	&13.8&	38.6&	51&	35.9&	17.7&	39.9&	51.6\tabularnewline
ScaleNet-101&37.3&	16.6&	42.4&	55.1&	39.5&	21.3&	44&	55.2\tabularnewline
\hline
\end{tabular}
\caption{Comparisons of mAP on MS COCO. mmAP indicates the results of mAP@IoU=[0.50:0.95]. Results of ResNets and ScaleNets are obtained by keeping all settings the same except backbone for fair comparison.  }
\label{tab:tab_coco}
\end{table*}

\textbf{ScaleNets are effective for object detection.} Table~\ref{tab:tab_effective_backbone} compares the effectiveness of backbones for object detection. It has been shown that ScaleNet-101 achieves the best detection performance with the minimal computational complexity amongst ResNets, ResNeXts~\cite{Xie2017}, SE-ResNets~\cite{Hu2018}, and Xception~\cite{Google2014}.

\begin{table}
\setlength{\abovecaptionskip}{4pt}
\begin{tabular}{c|cc|c}
\hline
 & \multicolumn{2}{c}{ImageNet} & COCO \tabularnewline
\hline
 & top-1 err. & FLOPs & mAP (600/1000)\tabularnewline
ResNet-152 & $21.58$ & $11.5$ & $34.3$\tabularnewline
Xception & $21.11$ & $9.0$ & $27.7$\tabularnewline
SE\_ResNet-152 & $21.07$ & $11.5$ & $37.1$\tabularnewline
ResNeXt-101 & $21.01$ & $8.0$ & $36.7$\tabularnewline
ScaleNet-101 & $20.97$ & \textbf{$\boldsymbol{7.5}$} & \textbf{$\boldsymbol{37.3}$}\tabularnewline
\hline
\end{tabular}
\caption{Comparisons of effectiveness of backbone for object detection on MS COCO.  All models are trained with the same strategy  for fair comparison. }
\label{tab:tab_effective_backbone}
\end{table}

\subsection{Analysis}

\textbf{The role of max pool.} Downsampling can be implemented in several ways: (\romannumeral1) a $3\times 3$ conv  with stride $s$;  (\romannumeral2) a dilated $3\times 3$ conv with stride $s$~\cite{Onvolutions2016}; (\romannumeral3)  a $s\times s$ avg pool with stride $s$; (\romannumeral4)   a $s\times s$ max pool with stride $s$. We evaluate all the above settings with ScaleNet-56 on CIFAR-100 by setting scale number $L$ to 2 and $s$ to 2 for simplicity. As shown in Table~\ref{tab:tab_down_up}, (\romannumeral4)  performs best. It suggests that max pool is the key factor of performance boosting. It is reasonable since  max pool preserves and enhances the maximum activation from previous layers so that the high response of small foreground regions would not be drowned by background features as information flows from bottom to top.

\begin{table}
\setlength{\abovecaptionskip}{4pt}
\center
\begin{tabular}{l|l}
\hline
method & top1 err.\tabularnewline
\hline
 stride 2 of $3\times 3$ conv & 26.19\tabularnewline
 stride 2 of $3\times 3$ conv, dilated 2 & 25.42\tabularnewline
 $2\times 2$ average pool & 24.58\tabularnewline
 $2\times 2$ max pool & \textbf{24.48}\tabularnewline
\hline
\end{tabular}
\caption{Top-1 error rate on CIFAR-100 with different downsampling methods. All the methods are of same FLOPs and record the best result in 5 runs.}
\label{tab:tab_down_up}
\end{table}

\begin{figure}[h]
\setlength{\abovecaptionskip}{4pt}
\tiny
\centering
\includegraphics[width=0.49\textwidth]{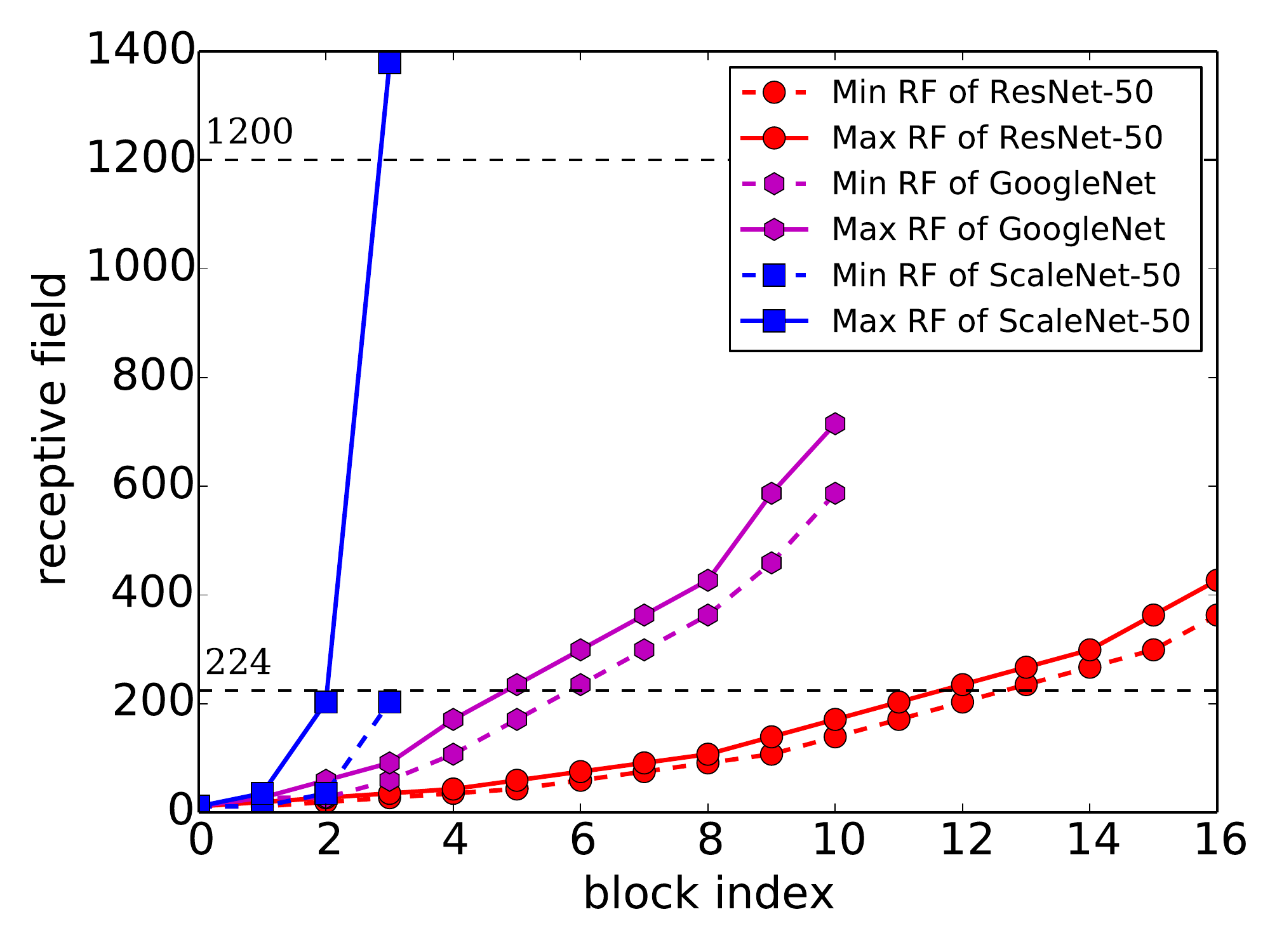}
\caption{Comparisons of receptive field of multi-branch networks as a function of block index. The shortcut branch and residual branch in each residual block of ResNets have the minimal and maximal receptive field respectively. The $1\times 1$ conv branch, and 5\texttimes 5 conv branch in each Inception block of GoogleNet have the minimal and maximal receptive field respectively. }
\label{fig:fig_rf_compare}
\end{figure}

\textbf{Wide range of receptive field.} Figure~\ref{fig:fig_rf_compare} compares the receptive field range of each block. It has been shown that the proposed ScaleNets have much wider range of receptive field than others.
Particularly, ScaleNet-50 reaches the resolution for classification and detection only in second and third block. On the one hand, ScaleNets potentially aggregate rich representations with large range of scales. On the other hand, they can extract global context information at very early stage (\eg, block 3) in one network. Together with data-driven neuron allocation, ScaleNets perform effectively and efficiently on various visual recognition tasks.

\section{Conclusion}
In this paper, we proposed a scale aggregation block with data-driven neuron allocation. The SA block can replace $3 \times 3$ conv in ResNets to get ScaleNets. The data-driven neuron allocation can effectively allocate the neurons to suitable scale in each SA block. The proposed ScaleNets have wide range of receptive fields, and perform effectively and efficiently on image classification and object detection. We will test our ScaleNets on more computer vision tasks, such as segmentation~\cite{Long2015}, in the future.

\section{Acknowledgement}
This work was supported by Beijing Municipal Science and Technology Commission (Z181100008918004) in part. We thank Zhanglin Peng, Youjiang Xu, Xinjiang Wang, and Huabin Zheng from SenseTime for the helpful suggestions and supports.

{\small
	\bibliographystyle{ieee}
	\bibliography{egbib}

\begin{thebibliography}{10}\itemsep=-1pt

\bibitem{channel_pruning}
https://github.com/yihui-he/channel-pruning.

\bibitem{Alvarez2016}
J.~M. Alvarez and M.~Salzmann.
\newblock Learning the number of neurons in neep networks.
\newblock In {\em NIPS}, pages 2270--2278. 2016.

\bibitem{Bell2015}
S.~Bell, C.~Lawrence~Zitnick, K.~Bala, and R.~Girshick.
\newblock Inside-outside net: Detecting objects in context with skip pooling
  and recurrent neural networks.
\newblock In {\em CVPR}, pages 2874--2883, 2016.

\bibitem{Cao2018}
X.~Cao, Z.~Wang, Y.~Zhao, and F.~Su.
\newblock Scale aggregation network for accurate and efficient crowd counting.
\newblock In {\em ECCV}, pages 734--750, 2018.

\bibitem{Google2014}
J.~Carreira, H.~Madeira, and J.~G. Silva.
\newblock Xception: A technique for the experimental evaluation of
  dependability in modern computers.
\newblock {\em IEEE Transactions on Software Engineering}, 24(2):125--136,
  1998.

\bibitem{Chen2018c}
X.~Chen.
\newblock Adaptive multi-scale information flow for object detection.
\newblock In {\em BMVC}.

\bibitem{chin2018layer}
T.-W. Chin, C.~Zhang, and D.~Marculescu.
\newblock Layer-compensated pruning for resource-constrained convolutional
  neural networks.
\newblock {\em arXiv preprint}, 2018.

\bibitem{Gordon}
A.~Gordon, E.~Eban, O.~Nachum, B.~Chen, H.~Wu, T.-J. Yang, and E.~Choi.
\newblock Morphnet: Fast \& simple resource-constrained structure learning of
  deep networks.
\newblock In {\em CVPR}, 2018.

\bibitem{he2016deep}
K.~He, X.~Zhang, S.~Ren, and J.~Sun.
\newblock Deep residual learning for image recognition.
\newblock {\em CVPR}, pages 770--778, 2016.

\bibitem{he2017channel}
Y.~He, X.~Zhang, and J.~Sun.
\newblock Channel pruning for accelerating very deep neural networks.
\newblock In {\em ICCV}, 2017.

\bibitem{Hu2018}
J.~Hu, L.~Shen, and G.~Sun.
\newblock Squeeze-and-excitation networks.
\newblock In {\em CVPR}, 2018.

\bibitem{huang2018multiscale}
G.~Huang, D.~Chen, T.~Li, F.~Wu, L.~van~der Maaten, and K.~Weinberger.
\newblock Multi-scale dense networks for resource efficient image
  classification.
\newblock In {\em ICLR}, 2018.

\bibitem{huang2016densely}
G.~Huang, Z.~Liu, L.~Van Der~Maaten, and K.~Q. Weinberger.
\newblock Densely connected convolutional networks.
\newblock In {\em CVPR}, volume~1, page~3, 2017.

\bibitem{huang2018data}
Z.~Huang and N.~Wang.
\newblock Data-driven sparse structure selection for deep neural networks.
\newblock In {\em ECCV}, pages 304--320, 2018.

\bibitem{ioffe2015batch}
S.~Ioffe and C.~Szegedy.
\newblock {Batch normalization: Accelerating deep network training by reducing
  internal covariate shift}.
\newblock In {\em ICML}, pages 448--456, 2015.

\bibitem{Kong2016}
T.~Kong, A.~Yao, Y.~Chen, and F.~Sun.
\newblock Hypernet: Towards accurate region proposal generation and joint
  object detection.
\newblock In {\em CVPR}, pages 845--853, 2016.

\bibitem{Krizhevsky2009}
A.~Krizhevsky and G.~Hinton.
\newblock Learning multiple layers of features from tiny images.
\newblock Technical report, 2009.

\bibitem{krizhevsky2012imagenet}
A.~Krizhevsky, I.~Sutskever, and G.~E. Hinton.
\newblock Imagenet classification with deep convolutional neural networks.
\newblock In {\em NIPS}, pages 1097--1105, 2012.

\bibitem{Lebedev2015a}
V.~Lebedev and V.~Lempitsky.
\newblock Fast convnets using group-wise brain damage.
\newblock In {\em CVPR}, pages 2554--2564, 2016.

\bibitem{Lin2017a}
T.-y. Lin, P.~Doll, R.~Girshick, K.~He, B.~Hariharan, S.~Belongie, F.~Ai, and
  C.~Tech.
\newblock Fpn feature pyramid networks for object detection.
\newblock {\em CVPR}, 2017.

\bibitem{Lin2014}
T.-Y. Lin, M.~Maire, S.~Belongie, J.~Hays, P.~Perona, D.~Ramanan,
  P.~Doll{\'a}r, and C.~L. Zitnick.
\newblock Microsoft coco: Common objects in context.
\newblock In {\em ECCV}, pages 740--755, 2014.

\bibitem{Long2015}
J.~Long, E.~Shelhamer, and T.~Darrell.
\newblock Fully convolutional networks for semantic segmentation.
\newblock In {\em CVPR}, pages 3431--3440, 2015.

\bibitem{Onvolutions2016}
D.~I.~C. Onvolutions.
\newblock {Multi-scale context aggregation by dilated convolutions}.
\newblock In {\em ICLR}, 2016.

\bibitem{Rabinovich2016}
A.~Rabinovich and A.~C. Berg.
\newblock Parsenet looking wider to see better.
\newblock pages 1--11, 2016.

\bibitem{Ren2017}
S.~Ren, K.~He, R.~Girshick, and J.~Sun.
\newblock {Faster R-CNN: towards real-time object detection with region
  proposal networks}.
\newblock {\em PAMI}, 39(6):1137--1149, 2017.

\bibitem{NIPS2016_6304}
S.~Saxena and J.~Verbeek.
\newblock Convolutional neural fabrics.
\newblock In D.~D. Lee, M.~Sugiyama, U.~V. Luxburg, I.~Guyon, and R.~Garnett,
  editors, {\em NIPS}, pages 4053--4061. Curran Associates, Inc., 2016.

\bibitem{Srivastava2015}
R.~K. Srivastava, K.~Greff, and J.~Schmidhuber.
\newblock Highway networks.
\newblock {\em arXiv preprint}, 2015.

\bibitem{Christian_cvpr2015}
C.~Szegedy, W.~Liu, Y.~Jia, P.~Sermanet, S.~Reed, D.~Anguelov, D.~Erhan,
  V.~Vanhoucke, and A.~Rabinovich.
\newblock Going deeper with convolutions.
\newblock In {\em CVPR}, pages 1--9, 2015.

\bibitem{Veniat2017}
T.~Veniat and L.~Denoyer.
\newblock Learning time/emory-efficient deep architectures with budgeted super
  networks.
\newblock {\em arXiv preprint}, 2017.

\bibitem{Williams1995}
P.~M. Williams.
\newblock Bayesian regularization and pruning using a laplace prior.
\newblock {\em Neural computation}, 7(1):117--143, 1995.

\bibitem{Xie2017}
S.~Xie, R.~Girshick, P.~Doll{\'a}r, Z.~Tu, and K.~He.
\newblock Aggregated residual transformations for deep neural networks.
\newblock In {\em CVPR}, pages 5987--5995. IEEE, 2017.

\bibitem{Yu2018}
F.~Yu, D.~Wang, E.~Shelhamer, and T.~Darrell.
\newblock Deep layer aggregation.
\newblock In {\em CVPR}, 2018.

\bibitem{yu2018nisp}
R.~Yu, A.~Li, C.-F. Chen, J.-H. Lai, V.~I. Morariu, X.~Han, M.~Gao, C.-Y. Lin,
  and L.~S. Davis.
\newblock Nisp: Pruning networks using neuron importance score propagation.
\newblock In {\em CVPR}, 2018.

\bibitem{Zhao2017}
H.~Zhao, J.~Shi, X.~Qi, X.~Wang, and J.~Jia.
\newblock Pyramid scene parsing network.
\newblock In {\em CVPR}, pages 2881--2890, 2017.

\bibitem{Zoph2017}
B.~Zoph and Q.~V. Le.
\newblock Neural architecture search with reinforcement learning.
\newblock {\em arXiv preprint}, 2016.

\end{thebibliography}
}

%\\{\small
%\bibliographystyle{ieee}
%\bibliography{egbib}
%}

%\end{document}

%\documentclass[10pt,twocolumn,letterpaper]{article}

%\usepackage{cvpr}
%\usepackage{times}
%\usepackage{epsfig}
%\usepackage{graphicx}
%\usepackage{amsmath}
%\usepackage{amssymb}
%\pagenumbering{gobble}
%\usepackage{color}
%\usepackage{array}
%\usepackage{float}
%\usepackage{textcomp}
%\usepackage{multirow}
%\usepackage{amsmath}
%\usepackage{amssymb}
%\usepackage{mathrsfs}
%\usepackage{commath}
%\usepackage{algorithm2e}
%\usepackage{caption} 
%\captionsetup{margin=1pt, font=small}
%\usepackage[section]{placeins}
%\usepackage{float}
%\captionsetup[table]{skip=10pt}
%\captionsetup{skip=9pt}

% Include other packages here, before hyperref.

% If you comment hyperref and then uncomment it, you should delete
% egpaper.aux before re-running latex.  (Or just hit 'q' on the first latex
% run, let it finish, and you should be clear).
%\usepackage[breaklinks=true,bookmarks=false]{hyperref}

\cvprfinalcopy % *** Uncomment this line for the final submission

\def\cvprPaperID{****} % *** Enter the CVPR Paper ID here
\def\httilde{\mbox{\tt\raisebox{-.5ex}{\symbol{126}}}}

% Pages are numbered in submission mode, and unnumbered in camera-ready
%\ifcvprfinal\pagestyle{empty}\fi

%\begin{document}
\newpage
\section*{Appendix A. Training Curves on ImageNet}
\begin{figure}[H]
%\center
\includegraphics[scale=0.44]{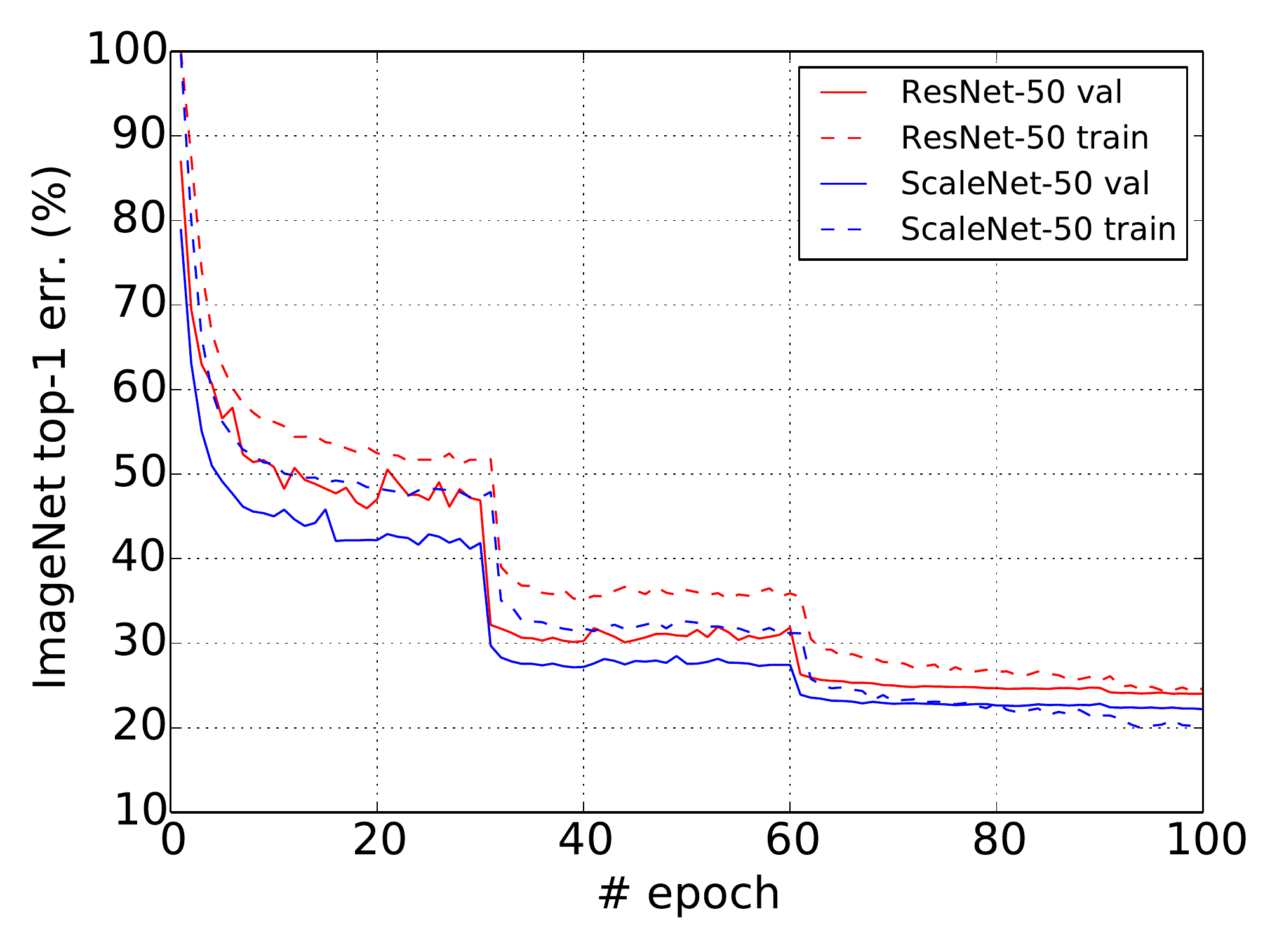}
\includegraphics[scale=0.44]{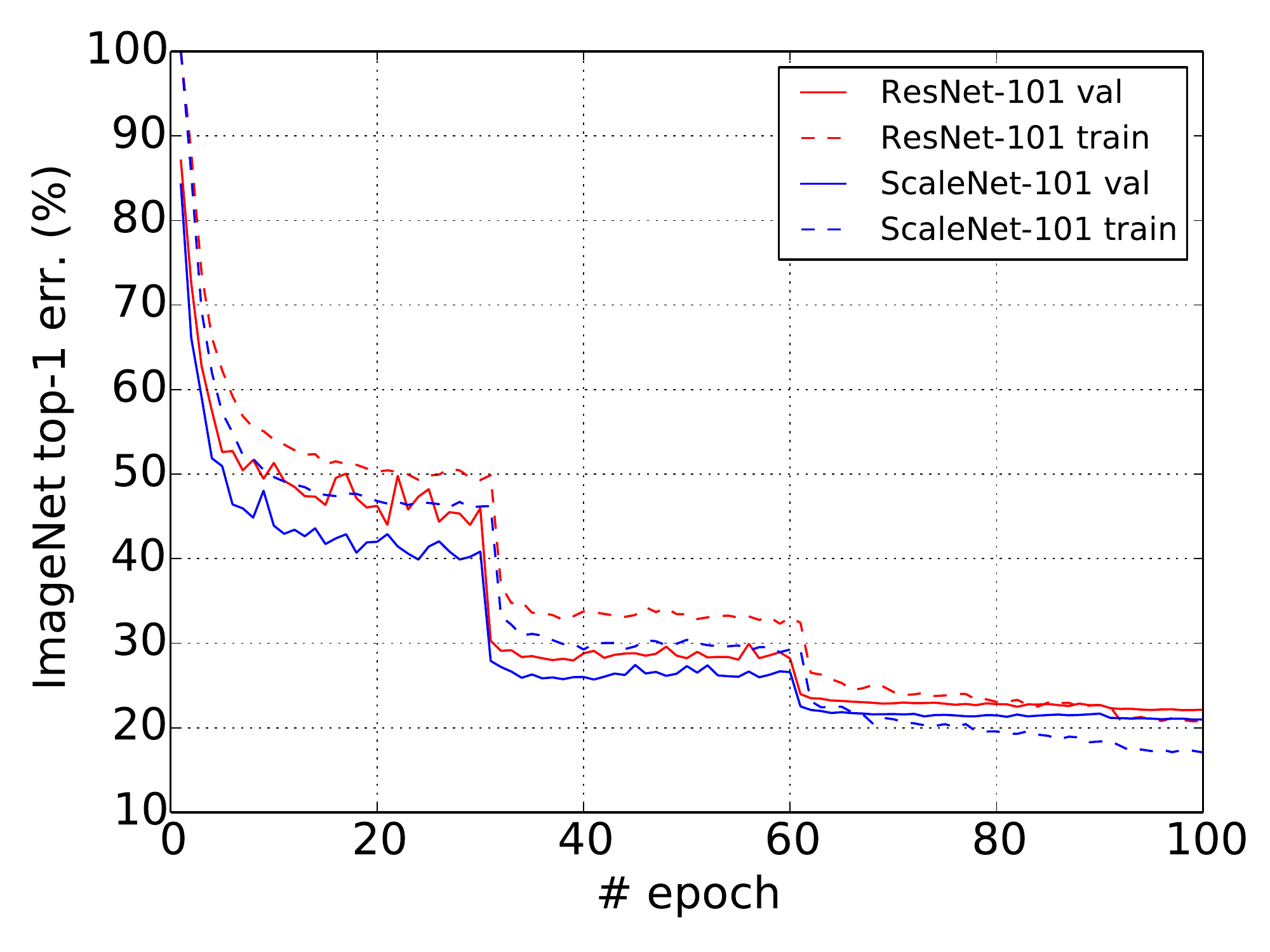}
\caption{Comparison of the training curves. The left compares scale ScaleNet-50
and ResNet-50 on ImageNet training set and validation set, while the
right compares ScaleNet-101 and ResNet-101.}
\label{fig::imagenet_error} 
\end{figure}

Figure \ref{fig::imagenet_error} compares the top-1 error rate of
ResNets and ScaleNets on both ImageNet training and validation dataset.
It has been shown that ScaleNets achieve much lower error rates than
their counterpart ResNets during all training phase.

\section*{Appendix B. Comparison with Pruning}
To demonstrate the advantages comparing with pruning, we list the stage-of-art pruning methods and ScaleNet-50-light in Table~\ref{tab::pruning}.
\begin{table}
\setlength{\abovecaptionskip}{4pt}
\center
\begin{tabular}{c|c|c}
\hline
 & top-1 acc.$\uparrow$& FLOPs (10$^9$)$\downarrow$\\ \hline
CP-ResNet-50~\cite{channel_pruning,he2017channel} & -3.68 & 1.5\\
SSS-ResNet-50~\cite{huang2018data} & -1.94 & 1.3\\
NISP-ResNet-50~\cite{yu2018nisp} & -0.21 & 1.1\\
LCP-ResNet-50~\cite{chin2018layer} & +0.09 & 1.0\\
ScaleNet-50-light & \textbf{+0.98} & 1.2\\
\hline
\end{tabular}\caption{Comparison with state-of-the-art methods on ImageNet. ResNet-50 and ScaleNet-50-light are trained in same settings, and others are reported in their papers.}
\label{tab::pruning}
\end{table}

\begin{table}
\setlength{\abovecaptionskip}{4pt}
\center
\begin{tabular}{c|c|c|c}
\hline
 & top-1 err. & FLOPs($19^9$) & GPU time(ms)\\
\hline
ResNet-50 & 24.02 & 4.1 & 95\\%95\\
\hline
SE-ResNet-50 & 23.29 & 4.1 & 98\\%98\\
\hline
ResNeXt-50 & \textbf{22.2} & 4.2 &147\\ %147\\
\hline
ScaleNet-50 & \textbf{22.2} & \textbf{3.8} & \textbf{93}\\%\textbf{93}\\
\hline
\end{tabular}
\caption{Comparison of GPU time (averaging on 1000 runs).}
\label{tab::time}
\end{table}

\section*{Appendix C. GPU time}
In Table~\ref{tab::time} all networks were tested using Tensorflow with GTX 1060 GPU and i7 CPU at batch size 16 and image size 224. It has been shown that ScaleNet-50 achieves the best accuracy on ImageNet while with the least GPU running time.

\section*{Appendix D. Allocated Neuron Numbers of ScaleNets}
Table~\ref{tab::neuron_number_detail} lists the detailed allocated neuron numbers of $3\times3$ conv in SA block on ImageNet and CIFAR-100.

{\small{}{}{}\clearpage}{ 
\begin{table*}
\center%
\begin{tabular}{c||c|c|c||c|c|c|c}
\hline 
\multirow{2}{*}{\multirow{2}{*}{block index} } & \multicolumn{3}{c||}{ScaleNets for CIFAR-100} & \multicolumn{4}{c}{ScaleNets for ImageNet}\tabularnewline
\cline{2-8} 
 & 38 layers & 56 layers  & 101 layers  & 50 layers (light)  & 50 layers  & 101 layers  & 152 layers\tabularnewline
\hline 
1  & 6,2,14  & 14,6,3  & 10,6,7  & 30,8,10,16  & 62,9,5,12  & 61,11,7,7  & 39,27,10,14\tabularnewline
\hline 
2  & 15,6,1  & 10,10,3  & 5,5,13  & 30,9,9,16  & 55,27,5,1  & 56,23,4,3  & 45,32,8,5\tabularnewline
\hline 
3  & 16,5,1  & 9,3,11  & 8,4,11  & 30,27,7,0  & 59,26,0,3  & 59,24,3,0  & 46,36,8,0\tabularnewline
\hline 
4  & 16,6,0  & 11,3,9  & 12,8,3  & 59,55,13,1  & 125,41,6,3  & 123,41,1,6  & 55,26,35,63\tabularnewline
\hline 
5  & 30,12,1  & 14,9,0  & 11,6,6  & 59,43,8,18  & 90,39,9,37  & 126,38,1,6  & 89,44,19,27\tabularnewline
\hline 
6  & 24,15,4  & 15,8,0  & 9,9,5  & 59,57,12,0  & 106,56,4,9  & 127,41,3,0  & 93,62,14,10\tabularnewline
\hline 
7  & 26,16,1  & 28,15,3  & 9,11,3  & 59,59,9,1  & 116,56,3,0  & 127,41,3,0  & 110,43,12,14\tabularnewline
\hline 
8  & 27,13,3  & 26,13,7  & 13,8,2  & 117,65,71,3  & 223,71,55,0  & 220,86,35,0  & 109,54,13,3\tabularnewline
\hline 
9  & 47,17,22  & 29,15,2  & 12,9,2  & 107,16,33,100  & 196,104,44,5  & 186,64,55,36  & 119,59,0,1\tabularnewline
\hline 
10  & 30,38,18  & 26,16,4  & 15,8,0  & 111,49,62,34  & 195,98,52,4  & 156,25,53,107  & 106,70,3,0\tabularnewline
\hline 
11  & 26,52,8  & 30,14,2  & 16,7,0  & 106,61,61,28  & 155,128,66,0  & 191,44,52,54  & 114,65,0,0\tabularnewline
\hline 
12  & 23,51,12  & 12,25,9  & 30,14,2  & 99,71,59,27  & 134,129,86,0  & 181,53,83,24  & 224,102,31,1\tabularnewline
\hline 
13  &  & 50,19,22  & 23,14,9  & 76,50,67,63  & 120,127,98,4  & 221,82,34,4  & 163,49,68,78\tabularnewline
\hline 
14  &  & 52,34,5  & 28,14,4  & 141,182,189,0  & 237,354,106,0  & 177,62,90,12  & 115,65,73,105\tabularnewline
\hline 
15  &  & 25,47,19  & 23,15,8  & 83,9,185,235  & 172,435,90,0  & 130,75,102,34  & 143,107,71,37\tabularnewline
\hline 
16  &  & 24,59,8  & 26,17,3  & 77,16,184,235  & 138,462,97,0  & 206,71,55,9  & 144,97,92,25\tabularnewline
\hline 
17  &  & 17,57,17  & 26,16,4  &  &  & 203,83,53,2  & 195,87,60,16\tabularnewline
\hline 
18  &  & 2,58,31  & 21,22,3  &  &  & 207,73,54,7  & 198,77,62,21\tabularnewline
\hline 
19  &  &  & 24,22,0  &  &  & 245,84,12,0  & 168,139,43,8\tabularnewline
\hline 
20  &  &  & 19,18,9  &  &  & 221,103,17,0  & 80,80,71,127\tabularnewline
\hline 
21  &  &  & 28,18,0  &  &  & 221,100,20,0  & 138,88,93,39\tabularnewline
\hline 
22  &  &  & 22,20,4  &  &  & 158,99,84,0  & 107,93,65,93\tabularnewline
\hline 
23  &  &  & 53,15,24  &  &  & 220,106,15,0  & 230,103,25,0\tabularnewline
\hline 
24  &  &  & 30,43,19  &  &  & 173,92,73,3  & 182,132,40,4\tabularnewline
\hline 
25  &  &  & 61,27,4  &  &  & 135,122,84,0  & 179,114,53,12\tabularnewline
\hline 
26  &  &  & 52,35,5  &  &  & 109,71,132,29  & 220,76,53,9\tabularnewline
\hline 
27  &  &  & 42,45,5  &  &  & 147,94,93,7  & 227,118,13,0\tabularnewline
\hline 
28  &  &  & 43,45,4  &  &  & 191,108,42,0  & 196,110,51,1\tabularnewline
\hline 
29  &  &  & 43,47,2  &  &  & 127,95,113,6  & 232,118,8,0\tabularnewline
\hline 
30  &  &  & 6,50,36  &  &  & 203,117,21,0  & 224,114,20,0\tabularnewline
\hline 
31  &  &  & 4,51,37  &  &  & 282,377,23,0  & 214,100,43,1\tabularnewline
\hline 
32  &  &  & 22,57,13  &  &  & 279,388,15,0  & 139,114,97,8\tabularnewline
\hline 
33  &  &  & 9,60,23  &  &  & 84,442,155,1  & 198,113,47,0\tabularnewline
\hline 
34  &  &  &  &  &  &  & 151,87,115,5\tabularnewline
\hline 
35  &  &  &  &  &  &  & 171,103,83,1\tabularnewline
\hline 
36  &  &  &  &  &  &  & 172,104,72,10\tabularnewline
\hline 
37  &  &  &  &  &  &  & 205,88,65,0\tabularnewline
\hline 
38  &  &  &  &  &  &  & 170,122,64,2\tabularnewline
\hline 
39  &  &  &  &  &  &  & 170,98,81,9\tabularnewline
\hline 
40  &  &  &  &  &  &  & 223,101,32,2\tabularnewline
\hline 
41  &  &  &  &  &  &  & 192,114,52,0\tabularnewline
\hline 
42  &  &  &  &  &  &  & 112,99,134,13\tabularnewline
\hline 
43  &  &  &  &  &  &  & 109,116,130,3\tabularnewline
\hline 
44  &  &  &  &  &  &  & 110,90,118,40\tabularnewline
\hline 
45  &  &  &  &  &  &  & 194,115,49,0\tabularnewline
\hline 
46  &  &  &  &  &  &  & 178,135,45,0\tabularnewline
\hline 
47  &  &  &  &  &  &  & 209,135,14,0\tabularnewline
\hline 
48  &  &  &  &  &  &  & 341,368,6,0\tabularnewline
\hline 
49  &  &  &  &  &  &  & 363,348,4,0\tabularnewline
\hline 
50  &  &  &  &  &  &  & 311,398,6,0\tabularnewline
\hline 
\end{tabular}

\caption{Learned neuron numbers in each SA block in ScaleNets on CIFAR-100 and ImageNet. These numbers indicates the output channel numbers for scale 1, 2, 3, and 4 (\eg $\text{3\texttimes3 conv}_{[C_{1},C_{2},C_{3},C_{4}]}$
in Table~\ref{tab:tab_archetecture}). Note that ScaleNets on CIFAR-100 have only three scales.}
\label{tab::neuron_number_detail}
\end{table*}}

\end{document}